\documentclass{article}

    \PassOptionsToPackage{numbers, compress}{natbib}

     \usepackage[final]{neurips_2019}

\usepackage[utf8]{inputenc} 
\usepackage[T1]{fontenc}    
\usepackage{hyperref}       
\usepackage{longtable}
\usepackage{tabu}
\usepackage{tabularx}
\usepackage{graphicx}
\usepackage{url}            
\usepackage{booktabs}       
\usepackage{amsfonts}       
\usepackage{nicefrac}       
\usepackage{microtype}      
\usepackage{amsmath}

\title{Modelling EHR timeseries by restricting feature interaction}

\author{%
  Kun Zhang\\
  Google Health \\
  \texttt{kunzhang@google.com} \\
  \And
  Yuan Xue\\
  Google Health \\
  \texttt{yuanxue@google.com} \\
  \And
  Gerardo Flores \\
  Google Health \\
  \texttt{gafm@google.com} \\
  \And
  Alvin Rajkomar \\
  Google Health \\
  \texttt{alvinrajkomar@google.com} \\
  \And
  Claire Cui \\
  Google Health \\
  \texttt{claire@google.com} \\
  \And
  Andrew M. Dai \\
  Google Health \\
  \texttt{adai@google.com} \\
}

\begin{document}

\maketitle

\begin{abstract}
Time series data are prevalent in electronic health records, mostly in the form of physiological parameters such as vital signs and lab tests. The patterns of these values may be significant indicators of patients' clinical states and there might be patterns that are unknown to clinicians but are highly predictive of some outcomes. Many of these values are also missing which makes it difficult to apply existing methods like decision trees. We propose a recurrent neural network model that reduces overfitting to noisy observations by limiting interactions between features. We analyze its performance on mortality, ICD-9 and AKI prediction from observational values on the Medical Information Mart for Intensive Care III (MIMIC-III) dataset. Our models result in an improvement of 1.1\% [p<0.01] in AU-ROC for mortality prediction under the MetaVision subset and 1.0\% and 2.2\% [p<0.01] respectively for mortality and AKI under the full MIMIC-III dataset compared to existing state-of-the-art interpolation, embedding and decay-based recurrent models.
\end{abstract}
\section{Introduction}

Observational values, such as lab results and vital signs, are frequently used to make a quantitative estimation of the current physiological state of a patient. However, these values are mostly processed into pre-specified ranges and buckets. For example, when calculating the commonly-used Acute Physiology and Chronic Health Evaluation (APACHE) IV score \citep{Zimmerman2006-xj}, there are as few as 3 buckets for some of the physiological measurements of the patients. These buckets have been assumed to be equally representative for all patients and ignore patients' different healthy baseline values. In addition, these score systems also ignore how the lab values are changing. For example, a systolic blood pressure that was rapidly trending from 111 to 219 would give the same NEWS score contribution of 0, although for many clinicians this would be an adverse indicator. These trend signals and many others are lost with many of the existing methods of processing lab values.

Predictive models such as mortality or billing code prediction utilise lab values, vital signs and other measurements to improve predictive accuracy. However, missing values are prevalent in EHR data since lab tests are ordered at the physician's discretion and costly or impractical measurements are not taken unless necessary. This results in time series data where the patterns of missingness can be predictive of risk or a diagnosis \citep{Schafer2002-ay}. For the modelling of time series data, observational values are typically standardized, while missing values are carried forward, interpolated from the previous value or are modelled to decay to the population mean \citep{che2016}. The patterns of missingness are typically represented as binary missingness indicator variables.

Recently, recurrent neural networks (RNNs) have been applied to electronic health records for more accurate clinical predictions \citep{Rajkomar2018-qm}. Overfitting is a common problem for deep learning models. Deep learning models are often overparameterized and so it is easy for the model to memorize the training data while failing to generalize to unseen data.

We introduce the feature-grouped long short-term memory network (FG-LSTM) that operates by modelling features individually and limiting their interactions in the model. The FG-LSTM specializes the long short-term memory network (LSTM \citep{Hochreiter1997-wc}) by restricting the form of the weight matrices. To ensure that missingness and time gaps are modelled, we represent each input feature (such as creatinine) by a group of two or three input variables: the standardized measurement value (interpolated if missing), a binary variable indicating presence or absence and an optional variable indicating the time since the feature was last measured. For a given input feature, the FG-LSTM allows all these components to interact but prevents features from interacting with each other. This reduces overfitting in the model and prevents learning spurious interactions or correlations between certain features over a few short timesteps. At inference time, features can only interact after each entire sequence of features has been read, which tends to produce smoother predictions over time.

\section{Methods}

In the FG-LSTM, each input feature is represented by a group of two or three variables (referred to as a feature group). We denote a multivariate time series indexed by $t$ as a vector $x_t=(u_t,v_t,w_t)$ where $u_t=(u_{1t},\ldots,u_{pt})$ denote the standardized values for the $p$ features, $v_t=(v_{1t},\ldots,v_{pt})$ denote the binary missing indicators where 0 indicates a feature is missing and $w_t=(w_{1t},\ldots,w_{pt})$ optionally denote the time since the last observation. The standardized value is linearly interpolated between adjacent values when it is missing (taking time into account), and simply carried forward when all future values are missing as in the interpolation baseline. The time differences are defined similarly to GRU-D where $s_t$ is the absolute time when the $t$\textsuperscript{th} observation was obtained (after windowing) and $s_1$ is set to 0. The time differences are normalised to be between 0 and 1.
\begin{align}
    w_{kt}=
\begin{cases}
s_t-s_{t-1} + w_{kt-1}, & t>1, v_{kt-1}=0\\
s_t-s_{t-1}, & t>1, v_{kt-1}=1\\
0, & t=1
\end{cases}
\end{align}
When the time differences are not used, the vector only consists of $u$ and $v$. $x_t$ represents the set of observations at timestep $t$. A naive setup would be to run $p$ small recurrent neural networks, one for each feature group, but running many small RNNs can potentially be inefficient due to not being able to use a single large matrix multiplication. Instead, we feed all $p$ feature groups into a single RNN where constrained weight matrices are used to restrict feature interaction. We describe this as a FG-LSTM (feature grouped long short-term memory network). We define FG-LSTM by the following equations (which are a variant of the LSTM equations):
\begin{align}
f_t&=\sigma((W_f\cdot M_w) x_t + (U_f\cdot M_u) h_{t-1} + b_f) \\
i_t&=\sigma((W_i\cdot M_w) x_t + (U_i\cdot M_u) h_{t-1} + b_i) \\
o_t&=\sigma((W_o\cdot M_w) x_t + (U_o\cdot M_u) h_{t-1} + b_o) \\
c_t&=f_t\cdot c_{t-1}+i_t\cdot \tanh((W_c \cdot M_w) x_t + (U_c\cdot M_u) h_{t-1} + b_c) \qquad h_t=o_t\cdot \tanh(c_t) 
\end{align}
Here, $\sigma$ denotes the sigmoid function, $\tanh$ denotes the hyperbolic tangent function, and $\cdot$ denotes the Hadamard (elementwise) product. The weights $W_{\{f, i, o, c\}}$, $U_{\{f, i, o, c\}}$ and bias terms $b_{\{f, i, o, c\}}$ are learned during training. $M_w$ is a fixed binary mask for the input-to-hidden weight matrices, and $M_u$ is a fixed binary mask for the hidden-to-hidden weight matrices. The effect of the mask is to restrict the weight matrix so that each element of the hidden state and cell state of the LSTM is computed from only one feature group. The mask is defined as follows.
\begin{align}
    M_{wij}=
\begin{cases}
1 & \text{if } i \mod p = j \mod p \\
0 & \text{otherwise}
\end{cases}
\end{align}
$M_u$ is defined similarly. The FG-LSTM can be considered similar to running $p$ individual LSTM models. The hidden state of the LSTM at the last timestep of the sequence is passed through a dense fully-connected layer to generate predictions. Only at this point are the activations of the layer computed from multiple features so that they can interact. A sigmoid or softmax activation is then applied depending on the task (sigmoid for binary (AKI/mortality), softmax for ICD-9). The model is trained to minimize the cross-entropy loss on the ground-truth labels. Models were optimized using AdaGrad \citep{Duchi:2011:ASM:1953048.2021068} or Adam \citep{Kingma2015-pp} depending on the model. Standard dropout techniques were applied to models including standard input and hidden-layer dropout\citep{Srivastava:2014:DSW:2627435.2670313}, variational input and hidden-layer dropout\citep{Gal:2016:TGA:3157096.3157211}, and zoneout\citep{krueger+al-2017-zoneout-iclr}.
 
For the baselines, we use the author provided Keras implementation of GRU-D, as well as standard median and linear interpolation. We have also reported the performance of FG-LSTM with and without the time difference.
In all experiments, 80\%, 10\% and 10\% of patients were used as the train, validation and test sets respectively randomly split based on patient ID. The validation set was used for model hyperparameter tuning for FG-LSTM and all the baselines through Gaussian process bandit hyperparameter optimization \citep{Desautels:2014:PET:2627435.2750368}.
The hyperparameter limits are listed in Table~\ref{tab:tuning_hyp} with the final tuned hyperparameters listed in Table~\ref{tab:tuned_hyp}. All models were implemented in TensorFlow \citep{Abadi2016-xa}.

\section{Dataset}

We conduct experiments on the MIMIC-III dataset \citep{Johnson2016-is}, a publicly available dataset of critical care records. Each patient's medical data during the first 48 hours in the current hospitalization is represented as a time series 
as described in \citet{Rajkomar2018-qm}.

Our cohort consists of inpatients hospitalized for at least 48 hours. We require the patient's age to be greater or equal to 18 years at the time of admission. We present results on the full cohort as well as MetaVision and CareVue subsets of the cohort which contain significantly different data as described in \citet{Mark2016}. The MetaVision cohort is the same cohort as used in \citet{che2016}, which has been claimed to be superior quality data. The test cohort is described in Table~\ref{tab:test-cohort}.


We use the top 100 observational features according to measurement frequency as predictor features; these are listed in Table~\ref{tab:features}. Each feature is standardized (transformed to have a median of 0 and standard deviation of 1) according to training set statistics.

Measurements are grouped into 20-minute windows, and we take the average if there are multiple measurements in the same window. A time step is skipped in the sequence if no features are present in that window. Outliers are handled by clipping the value to 10 standard deviations. 

\section{Experiments} 
The following outcomes are predicted for each patient using the predictor variables described above. All predictions are at 48 hours after admission. More details are in the appendix.

\textbf{Mortality}: Whether the patient dies during the current hospital admission.

\textbf{AKI}: Predicting acute kidney injury (AKI) onset within the inpatient encounter.

\textbf{ICD-9 20 task classification}: The ICD-9 diagnosis codes are grouped into 20 categories following \citet{che2016} .

We report results with our model (FG-LSTM) along with several baselines. All baselines concatenate the input with a missingness indicator for each feature (unless mentioned) and use a LSTM model (unless specified). Outliers are handled by clipping the value to 10 standard deviations. Our preliminary experiments indicate that these outliers carry information and that removing them from the data results in a loss of performance. The baselines we use are described in detail in the appendix.

We report the test-set performance over 5 runs using the best validation set hyperparameters from different random initializations. We report both the area under the receiver operating characteristic curve (AU-ROC) and the area under the precision recall curve (AU-PRC).

Table~\ref{tab:full-results} compares the performance of our model (FG-LSTM) with several state of the art baselines on the full MIMIC-III dataset. The FG-LSTM results in significant absolute increases in AU-ROC of 1.0\% (Welch's t-test: P<0.001) and 2.2\% (Welch's t-test: P<0.0001) respectively for mortality and AKI compared to the best baseline models (interpolation and GRU-D). For the task of ICD-9 20 task classification we find our model's results are not significantly different from that of the GRU-D. We find that for ICD9 20 task classification, using the time differences improves performance, whereas there is no significant impact for mortality and AKI classification. We show the results of further ablations with FG-LSTM in Table~\ref{tab:ablation}.
\begin{table}
  \centering 
  \caption{Results on patient mortality, AKI and ICD-9 20 task classification at 48 hours after admission on the MIMIC-III dataset. We report the mean (standard deviation) for each metric over five repeated runs. We also report the significance of the difference between the FG-LSTM results and the best baseline model under Welch's t-test, where applicable.} 
  \label{tab:full-results} 
  \begin{tabularx}{\textwidth}{p{3cm}XXXX}\toprule
    & \multicolumn{2}{c}{Mortality} & \multicolumn{2}{c}{AKI} \\ 
    & \multicolumn{1}{c}{AU-ROC} & \multicolumn{1}{c}{AU-PRC} & \multicolumn{1}{c}{AU-ROC} & \multicolumn{1}{c}{AU-PRC} \\ \cmidrule(l){2-5}
    Percentile embedding w/o indicator & 0.8344 (0.0015) & 0.3456 (0.0081) & 0.7159 (0.0058) & 0.4297 (0.0095) \\
    Percentile embedding & 0.8371 (0.0024) & 0.3437 (0.0062) & 0.7205 (0.0050) & 0.4365 (0.0054) \\
    Median & 0.8399 (0.0021) & 0.3864 (0.0094) & 0.7316 (0.0031) & 0.4501 (0.0070) \\
    Interpolation & 0.8564 (0.0032) & 0.4009 (0.0122) & 0.7433 (0.0021) & 0.4630 (0.0047) \\
    GRU-D & 0.8544 (0.0033) & 0.4195 (0.0084) & 0.7474 (0.0025) & 0.4688 (0.0050) \\
    FG-LSTM & \textbf{0.8665 (0.0020)}*** & \textbf{0.4225 (0.0065)} & \textbf{0.7689 (0.0023)}*** & \textbf{0.4785 (0.0036)}** \\
    FG-LSTM w/ time differences & 0.8630 (0.0030) & 0.4126 (0.0033) & 0.7489 (0.0022) & 0.4679 (0.0055) \\ \midrule
    & \multicolumn{2}{c}{ICD-9 20 task classification} & & \\ \cmidrule(lr){2-3}
    Percentile embedding w/o indicator & 0.8444 (0.0004) & 0.7408 (0.0005) & & \\
    Percentile embedding & 0.8465 (0.0006) & 0.7450 (0.0009) & & \\
    Median & 0.8495 (0.0003) & \textbf{0.7515 (0.0005)} & & \\
    Interpolation & 0.8492 (0.0004) & 0.7500 (0.0004) & & \\
    GRU-D & 0.8489 (0.0004) & 0.7506 (0.0008) & & \\
    FG-LSTM & 0.8488 (0.0003) & 0.7500 (0.0003) & & \\
    FG-LSTM w/ time differences & \textbf{0.8496 (0.0003)} & 0.7511 (0.0005) & & \\ \bottomrule
  \end{tabularx}
  \caption*{**p < 0.01
    ***p < 0.001}
  
\end{table}

We also conducted experiments on admissions restricted to patients monitored using the MetaVision system in MIMIC-III. This is similar to the cohort from the GRU-D paper \citep{che2016}. Table~\ref{tab:metavision} compares the performance of FG-LSTM and the baselines trained and tested under the MetaVision subset of the dataset, which is claimed by \citet{che2016} to be superior quality in terms of time series data. We see a drop in performance on this subset, likely because it is only a third of the size of the full dataset. The FG-LSTM results in a significant absolute improvement of 1.1\% (Welch's t-test: P=0.0081) in AU-ROC for mortality under this MetaVision subset. Again for the task of ICD-9 20 task classification, there is no significant difference from GRU-D.
\section{Discussion} 
Our results show that the FG-LSTM performs significantly (under Welch's t-test) better than the state-of-the-art baseline methods (GRU-D and linear feature interpolation) for mortality and AKI prediction. These tasks are particularly sensitive to vital signs and lab values so it's reasonable that the FG-LSTM models these well. The insignificant results on ICD-9 20 class prediction is likely because the input features we chose were not significantly predictive of different diagnoses and it is possible that the other categorical or notes data in the EHR are better predictors for this task. In the appendix, we show that the FG-LSTM also yields more interpretable attribution than the baseline models. For future work, we expect the combination of the FG-LSTM with a model that handles categorical features as in \citet{Rajkomar2018-qm} can lead to better predictions for diagnosis, mortality and AKI.

\bibliography{references}
\bibliographystyle{unsrtnat}

\appendix
\onecolumn
\section*{Appendix}
\setcounter{table}{0}
\renewcommand{\thetable}{S\arabic{table}}
\begin{table}
  \centering 
  \caption{Results on patient mortality and the ICD-9 20 task at 48 hours after admission after training and testing on the MetaVision subset of MIMIC-III.} 
  \label{tab:metavision}
  \begin{tabularx}{\textwidth}{XXXXX}\toprule
    & \multicolumn{2}{c}{Mortality} & \multicolumn{2}{c}{ICD-9 20 task classification} \\
    & \multicolumn{1}{c}{AU-ROC} & \multicolumn{1}{c}{AU-PRC} & \multicolumn{1}{c}{AU-ROC} & \multicolumn{1}{c}{AU-PRC} \\ \cmidrule(l){2-5}
    Percentile embedding w/o indicator & 
0.8218 (0.0038) &
0.2903 (0.0030) &
0.8384 (0.0007) &
0.7726 (0.0012)
 \\
    Percentile embedding & 
0.8378 (0.0058) &
0.3267 (0.0068) &
0.8374 (0.0007) &
0.7716 (0.0011)
 \\
    Median & 
0.8417 (0.0043) &
0.3677 (0.0176) &
0.8379 (0.0005) &
0.7727 (0.0008)
 \\
    Interpolation & 
0.8373 (0.0099) &
0.3473 (0.0267) & 
0.8402 (0.0003) &
0.7762 (0.0003)
 \\
    GRU-D & 
0.8484 (0.0037) &
\textbf{0.3856 (0.0057)} &
0.8410 (0.0005) &
0.7787 (0.0007)
 \\
    FG-LSTM & 
\textbf{0.8591 (0.0054)}** &
0.3757 (0.0101) &
0.8419 (0.0002) &
\textbf{0.7796 (0.0003)}
 \\
    FG-LSTM w/ time differences & 
0.8567 (0.0019) &
0.3813 (0.0087) &
\textbf{0.8420 (0.0005)} &
0.7793 (0.0005)
 \\ \bottomrule
  \end{tabularx}
  \caption*{**p < 0.01}
\end{table}
\subsection*{Task details} 

The following outcomes are predicted for each patient using the predictor variables described above.
\begin{description}

\item[Mortality] Whether the patient dies during the current hospital admission. Predicted at 48 hours after admission. The dataset contains 46,120 admission records from 35,440 patients, with 4,277 positive labels.

\item[AKI] Predicting acute kidney injury (AKI) onset within the inpatient encounter at 48 hours after admission. This dataset contains 46,120 records from 35,440 patients, and has 10,180 positive labels.

AKI is a sudden episode of kidney failure or kidney damage that happens within a few hours or a few days. It is a common complication among hospitalized patients, and is an important cause for in-hospital death. Multiple criteria exist for AKI diagnosis. We adopt the KDIGO (Kidney Disease Improving Global Organization) criteria based on short-term lab value changes in our prediction tasks here:
\begin{itemize}
    \item Increase in serum creatinine by $\geq$ 0.3 mg/dl ($\geq$ 26.5 umol/l) within 48 hours;
    \item Urine volume < 0.5 ml/kg/h (25ml/h, assuming 50kg weight) for 6 hours.
\end{itemize}
At 48 hours after admission, we classify the patients who have not developed AKI but will have AKI within this encounter as positive and the others as negative examples.

\item[ICD-9 20 task classification] The ICD-9 diagnosis codes are grouped into 20 categories following \citet{che2016} . This is then predicted at 48 hours after admission, which has a total of 46,120 admission records from 35,440 patients.
\end{description}
\subsection*{Baselines}
\begin{description}
\item[Percentile embedding w/o indicator] Features are bucketed by percentiles and then the buckets are embedded, where each bucket embedding is initialized to a random vector and trained jointly. Any missing values are ignored. The number of buckets is tuned on the validation set. This is the method as described in the deep models in \citet{Rajkomar2018-qm}.
\item[Percentile embedding] As in the model above but the embedding vector is concatenated with a missingness indicator for each feature.
\item[Median] Standardized feature values are used and missing values are filled in with the median from the training set.
\item[Interpolation] Standardized feature values are used and linear interpolation is used to fill in the missing values. To interpolate a missing value $v$ at time $t$ between 2 measurements $v_1$ measured at $t_1$ and $v_2$ measured at $t_2$, $v=v_1+(v_2-v_1) \frac{t - t_1}{t_2 - t_1}$. If there is no measurement after $v_1$, the value $v_1$ is simply carried forward. If there is no measurement before $v_2$, the value $v_2$ is carried backward. If there is no measurement during the period, 0 will be used.
\item[GRU-D] The GRU based model as implemented by \citet{che2016} in TensorFlow which has trainable decay rates for the input and hidden states.
\end{description}
\subsection*{Attribution Methods}
Deep learning techniques are typically regarded as black boxes where it is hard to determine what causes a model to make a prediction. Recent advances in interpretability techniques have produced better tools to probe a trained model. One of these is path-integrated gradients \citep{Sundararajan2017-mg}. Gradients can be used to approximate the change in a prediction given a step change in the input data. Path-integrated gradients have been shown to produce a better approximation of the change in a prediction by summing gradients over a gradual change in the input data. This has typically been applied to images but here we adapt it to time series data.

To apply this technique to sparsely measured time series for a particular patient, we use as a baseline a patient who has had the same measurements recorded at the same times, but for whom all measurements take the population median value.  We then average the gradients of the model prediction across 50 evenly-spaced points between this baseline and the actual measurements.  For each lab and measurement time, we take the product of this averaged gradient with the change from measurement to baseline value as a linearized approximation of the influence of that value on the generated prediction.  Because the population median is mapped to zero in our normalization, we can represent the contribution of each lab type and time of measurement simply.  If $F(x)$ is the neural network's predicted probability of an event, as a function of the first 48 hours of lab values, then:
\begin{align}
    \textrm{IntGrad}(x_{it}) = \frac{x_{it}}{50}\sum_{k=1}^{50} \frac{\partial F(kx/50)}{\partial x_{it}}
\end{align}
  \subsection*{Dataset details}
  \renewcommand*{\arraystretch}{1.1}
  \begin{longtabu}{p{5cm}XXXX} \caption{Descriptive statistics for patient cohort. These consist of inpatients who are admitted for at least 48 hours in the MIMIC-III dataset and are used for training or validation purposes.} \\ \toprule
Demographics & \multicolumn{2}{c}{Adult MIMIC admissions} & \multicolumn{2}{p{3cm}}{MetaVision Only (GRU-D cohort)} \\ \midrule 
\endhead
\bottomrule
\endfoot
Number of Patients&
31,786& &
14,467 & \\
Number of Encounters &
41,387 & &
17,777 & \\
Number of Female Patients &
18,210 &
44.2\% &
7,874 &
44.3\% \\
Median Age
(Interquartile Range) &
66 &
 (25) &
66 &
(24) \\ \midrule
Disease Cohort & & & & \\ \cmidrule(r){1-1}
Cancer &
2,978 &
7.2\% &
1,359 &
7.6\% \\
Cardiopulmonary &
4,279 &
10.3\% &
1,924 &
10.8\%\\
Cardiovascular &
10,515 &
25.4\% &
3,715 &
20.9\% \\
Medical &
17,862 &
43.2\% &
8,409 &
47.3\% \\
Neurology &
4,998 &
12.1\% &
2,218 &
12.5\% \\
Obstetrics &
131 &
0.3\% &
47&
0.3\% \\
Psychiatric &
28 &
0.1\%&
18&
0.1\%\\
Other&
596&
1.4\%&
87&
0.5\%\\ \midrule
Number of Previous Hospitalizations&&&&\\ \cmidrule(r){1-1}
0&
31,463&
76.0\%&
12,874&
72.4\%\\
1&
5,932&
14.3\%&
2,725&
15.3\%\\
2-5&
3,429&
8.3\%&
1,845&
10.4\%\\
6+&
563&
1.4\%&
333&
1.9\%\\ \midrule
Discharge Disposition&&&&\\ \cmidrule(r){1-1}
Expired&
3,858&
9.3\%&
1,520&
8.6\%\\
Home&
21,022&
50.8\%&
8,876&
49.9\%\\
Other&
1,079&
2.6\%&
599&
3.4\%\\
Other Healthcare Facility&
2,938&
7.1\%&
1,809&
10.2\%\\
Rehabilitation&
5,706&
13.8\%&
1,657&
9.3\%\\
Skilled Nursing Facility&
6,784&
16.4\%&
3,316&
18.7\%\\ \midrule
Binary Label Prevalence&&&&\\ \cmidrule(r){1-1}
Mortality&
3,858&
9.3\%&
1,520&
8.6\%\\
Acute Kidney Injury (AKI)&
9,110&
22.0\% & & \\ \midrule
Multilabel Prevalence (ICD9 Groups)&&&&\\ \cmidrule(r){1-1}
1:Infectious and
Parasitic Diseases &
11,632&
28.1\%&
5,827&
32.8\%\\
2:Neoplasms&
7,293&
17.6\%&
3,651&
20.5\%\\
3:Endocrine, Nutritional and Metabolic Diseases, Immunity&
28,749&
69.5\%&
13,929&
78.4\%\\
4:Blood and
Blood-Forming Organs&
15,732&
38.0\%&
8,340&
46.9\%\\
5:Mental Disorders&
13,232&
32.0\%&
7,558&
42.5\%\\
6:Nervous System
and Sense Organs&
12,913&
31.2\%&
8,004&
45.0\%\\
7:Circulatory System&
34,985&
84.5\%&
15,327&
86.2\%\\
8:Respiratory System&
20,422&
49.3\%&
9,379&
52.8\%\\
9:Digestive System&
17,289&
41.8\%&
8,735&
49.1\%\\
10:Genitourinary System&
17,947&
43.4\%&
9,066&
51.0\%\\
11:Complications of Pregnancy, Childbirth,
and the Puerperium&
142&
0.3\%&
52&
0.3\%\\
12:Skin and
Subcutaneous Tissue&
4,852&
11.7\%&
2,494&
14.0\%\\
13:Musculoskeletal System and Connective Tissue&
8,349&
20.2\%&
4,929&
27.7\%\\
14:Congenital Anomalies&
1,442&
3.5\%&
725&
4.1\%\\
15:Symptoms&
12,979&
31.4\%&
7,506&
42.2\%\\
16:Nonspecific Abnormal Findings&
3,786&
9.2\%&
2,176&
12.2\%\\
17:Ill-defined and Unknown Causes of Morbidity and Mortality&
1,364&
3.3\%&
955&
5.4\%\\
18:Injury and Poisoning&
18,211&
44.0\%&
8,044&
45.2\%\\
19:Supplemental V-Codes&
21,607&
52.2\%&
12,001&
67.5\%\\
20:Supplemental E-Codes&
13,512&
32.7\%&
7,608&
42.8\%\\
  \end{longtabu}
  
  \subsection*{Results on MIMIC-III CareVue subset} 
We also took the model trained on the full MIMIC-III cohort and analysed results solely on the CareVue subset (the records not in the MetaVision cohort) to determine if the quality of data affected the relative performance of the models.

Table~\ref{tab:carevue} compares the performance of FG-LSTM and the baselines under the CareVue subset of the dataset, which are not considered by \citet{che2016} as they claim the data is worse quality. Again, for this subset we see a significant improvement in performance in mortality and AKI prediction using the FG-LSTM model as compared to the baselines. Again for the task of ICD-9 20 task classification, there is no significant difference from GRU-D. Interestingly, this shows that the poorer quality data does not affect the relative performance of the model.

\begin{table}
  \centering 
  \caption{Results on patient mortality, AKI and ICD-9 20 task at 48 hours after admission on the CareVue subset of MIMIC-III.} \label{tab:carevue}
  \begin{tabularx}{\textwidth}{p{3cm}XXXX}\toprule
    & \multicolumn{2}{c}{Mortality} & \multicolumn{2}{c}{AKI} \\ 
    & \multicolumn{1}{c}{AU-ROC} & \multicolumn{1}{c}{AU-PRC} & \multicolumn{1}{c}{AU-ROC} & \multicolumn{1}{c}{AU-PRC} \\ \cmidrule(l){2-5}
    Percentile embedding w/o indicator & 
0.8280 (0.0027)&
0.3456 (0.0094)&
0.7078 (0.0063)&
0.4250 (0.0099)
 \\
    Percentile embedding & 
0.8298 (0.0024)&
0.3488 (0.0036)&
0.7152 (0.0052)&
0.4371 (0.0040)
 \\
    Median & 
0.8318 (0.0044)&
0.3878 (0.0102)&
0.7243 (0.0051)&
0.4472 (0.0077)
 \\
    Interpolation & 
0.8516 (0.0025)&
0.4090 (0.0118)&
0.7407 (0.0010)&
0.4675 (0.0038)
 \\
    GRU-D & 
0.8560 (0.0032)&
\textbf{0.4380 (0.0084)}&
0.7417 (0.0019)&
0.4656 (0.0034)
 \\
    FG-LSTM & 
\textbf{0.8659 (0.0016)}***&
0.4362 (0.0111)&
\textbf{0.7691 (0.0023)}***&
\textbf{0.4843 (0.0048)}***
 \\
    FG-LSTM w/ time differences & 
0.8620 (0.0020)&
0.4251 (0.0076)&
0.7445 (0.0046)&
0.4683 (0.0078)
 \\ \midrule
    & \multicolumn{2}{c}{ICD-9 20 task classification} & & \\ \cmidrule{2-3}
    Percentile embedding w/o indicator & 
0.8441 (0.0006)&
0.7068 (0.0011)
 & & \\
    Percentile embedding & 
0.8463 (0.0006)&
0.7124 (0.0010)
 & & \\
    Median & 
0.8497 (0.0003)&
0.7195 (0.0003)
 & & \\
    Interpolation & 
0.8503 (0.0005)&
0.7202 (0.0006)
 & & \\
    GRU-D & 
0.8495 (0.0004)&
0.7194 (0.0010)
 & & \\
    FG-LSTM & 
0.8499 (0.0005)&
0.7197 (0.0008)
 & & \\
    FG-LSTM w/ time differences & 
\textbf{0.8510 (0.0005)}&
\textbf{0.7208 (0.0010)}
& & \\ \bottomrule
  \end{tabularx}
  \caption*{***p < 0.001}
\end{table}

\subsection*{Attribution}
\begin{figure}
  \centering 
  \includegraphics[width=0.7\columnwidth]{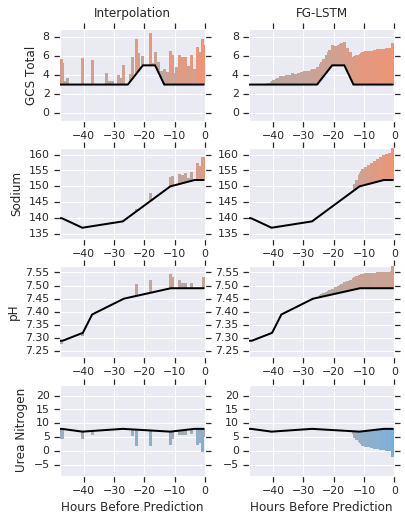} 
  \caption{We show 4 features that contribute to the FG-LSTM's prediction (right) in a different way than to the baseline interpolation model (left) via attribution, and their values for the preceding 48 hours prior to the prediction.  The red-overlays indicate that the particular value had a positive attribution to the predicted risk for that model and the height of the red columns are proportional to the log-scaled attribution weight.  The blue overlays indicate a negative attribution, which indicate a negative contribution to predicted risk for that model from that feature. GCS is the Glascow Coma Scale.}
  \label{fig:attribution} 
\end{figure} 

Figure~\ref{fig:attribution} shows attribution over time for a particular patient to the four lab measurements with the biggest difference in attribution between the interpolation model and the FG-LSTM. The patient had a persistently low Glascow Coma Score (GCS) for the 48 hours preceding the prediction, which indicates that the patient had poor neurological function. The sodium and pH values indicate progressive hypernatremia and alkalosis, which are clinically considered to represent a worsening physiological state. The blood urea nitrogen level remained constant, which clinically correlates with stable kidney functions. The FG-LSTM and interpolation model have directionally similar attributions (in line with clinical expectations), but FG-LSTM's attributions are more stable and smooth whereas the interpolation model has abrupt jumps in attribution despite small or no changes in the feature value. This is likely due to the interpolation model being overly sensitive to combinations of feature values over short periods of time. This can result in abrupt changes to predicted risk as measurements come in to the interpolation model as compared to the FG-LSTM.

\subsection*{Model Selection}
We tuned the models with the following hyperparameters, targeting AU-ROC on the full MIMIC-III dataset. For the LSTM based models, the AdaGrad optimizer was used, for the GRU-D model, the Adam optimizer was used with batchnorm as in the paper. For the regularization techniques used, i.e. input dropout, LSTM hidden state dropout, projection layer dropout, zoneout, and variational dropout, we use $P_k$ to denote keep probability, which is 1 - dropout probability.

\begin{table}
  \centering 
  \caption{Hyperparameter tuning limits used.} \label{tab:tuning_hyp}
  \begin{tabularx}{\textwidth}{p{6cm}XX}\toprule
  Hyperparameter & Minimum & Maximum \\ \midrule
Clip norm&
0.1&
50.0\\
Input dropout $P_k$&
0.01&
1.0\\
RNN Hidden dropout $P_k$&
0.01&
1.0 \\
Learning rate&
0.0001&
0.5\\
Percentile embedding size&
25&
200\\
Number of percentile buckets&5&20\\
RNN hidden size&
16&
3000 \\
RNN hidden size per feature group&
1&
30 \\
Projection layer dropout $P_k$&
0.01&
1.0\\
Projection layer size &
0&
1000\\
Variational input $P_k$ &
0.01&
1.0\\
Variational output $P_k$&
0.01&
1.0\\
Variational recurrent $P_k$&
0.01&
1.0\\
Zoneout $P_k$&
0.01&
1.0\\ \bottomrule
  \end{tabularx}
\end{table}
\begin{table}
  \centering 
  \caption{Tuned hyperparameters found through Gaussian process bandit optimization.} \label{tab:tuned_hyp}
  \begin{tabularx}{\textwidth}{p{2.5cm}XXXXX}\toprule
  Hyperparameter & Median & Percentile embedding & Interpolation & GRU-D & FG-LSTM \\ \midrule
Clip norm&
46.9164&
48.9696&
8.06007&
32.90225&
42.327009\\
Input dropout $P_k$& 0.487627& 0.326633 & 0.668343 & 0.747041 & 0.982881\\
RNN Hidden dropout $P_k$& 0.854115 &0.701658& 0.88545 &0.976599  & 0.356855 \\
Learning rate& 0.124912  & 0.19047 & 0.135474 & 0.001279 & 0.051977 \\
Percentile embedding size& N/A &126 & N/A& N/A&N/A \\
Number of percentile buckets & N/A & 4 &N/A& N/A&N/A \\
RNN hidden size& 114 & 73 & 309 & 187 & N/A \\
RNN hidden size per feature group & N/A &N/A & N/A & N/A & 21 \\
Projection layer dropout $P_k$& 0.888716 & 0.874535 & 0.973923 & 0.987385 & 0.992444 \\
Projection layer size &380 & 274 & 951 & 191 & 477 \\
Variational input $P_k$ & 0.951351 & 0.491936 & 0.992106 &N/A &N/A \\
Variational output $P_k$&0.990069 & 0.980551 & 0.856734 &N/A  &N/A \\
Variational recurrent $P_k$& 0.974393 & 0.979701 & 0.643025 & 0.970241 & 0.986196 \\
Zoneout $P_k$& 0.358289 & 0.989134 & 0.748179 &N/A  & 0.582535 \\
\bottomrule
  \end{tabularx}
\end{table}

\subsection*{Model ablation}
We also conducted a few ablation experiments on the FG-LSTM on the mortality task.
\begin{description}
\item[W/o indicator] missingness indicators are removed from input.
\item[W/o interpolation] missing values are filled with the median instead of interpolation.
\end{description}
All ablation experiments showed a significant drop of performance.
\begin{table}
  \centering 
  \caption{FG-LSTM model ablation experiments results on mortality dataset. We report the mean (standard deviation) for each metric over five repeated runs.} 
  \label{tab:ablation}
  \begin{tabularx}{\textwidth}{XXX}\toprule
& AU-ROC& AU-PRC \\ \cmidrule(l){2-3}
Full FG-LSTM&
0.8665 (0.0020)&
0.4225 (0.0065)\\
w/o indicator&
0.8576 (0.0020)&
0.4215 (0.0044)\\
w/o interpolation&
0.8494 (0.0032)&
0.3724 (0.0022)\\
w/o indicator and w/o interpolation&
0.8420 (0.0010)&
0.3674 (0.0087)\\ \bottomrule
\end{tabularx}
\end{table}

\begin{table}
  \centering 
  \caption{Kernel size (total size of $W_{\{f, i, o, c\}}, U_{\{f, i, o, c\}}$) comparison between the baseline interpolation model and the proposed FG-LSTM model.} \label{tab:kernel-size}
  \begin{tabularx}{\textwidth}{XXXX}\toprule
Hidden state size&
Corresponding
FG-LSTM 
Per feature group state size&
Size of baseline LSTM kernel&
Effective Size of FG-LSTM kernel\\ \midrule
50&
--&
50,000&
--\\
100&
1&
120,000&
1,200\\
200&
2&
320,000&
3,200\\
300&
3&
600,000&
6,000\\
400&
4&
960,000&
9,600\\
500&
5&
1,400,000&
14,000\\
1000&
10&
4,800,000&
48,000\\
1500&
15&
10,200,000&
102,000\\
2000&
20&
17,600,000&
176,000\\
\bottomrule
\end{tabularx}
\end{table}

  \renewcommand*{\arraystretch}{1.1}
  \begin{longtabu}{p{5cm}XXXX} \caption{Descriptive statistics for patient cohort in test set.} \label{tab:test-cohort} \\ \toprule
Demographics & \multicolumn{2}{c}{Adult MIMIC admissions} & \multicolumn{2}{p{3cm}}{MetaVision Only (GRU-D cohort)} \\ \midrule 
\endhead
\bottomrule
\endfoot
Number of Patients&
3,654& &
1,655 & \\
Number of Encounters &
4,733 & &
2,042 & \\
Number of Female Patients &
1,970 &
41.6\% &
872 &
42.7\% \\
Median Age
(Interquartile Range) &
66 &
 (24) &
66 &
(25) \\ \midrule
Disease Cohort & & & & \\ \cmidrule(r){1-1}
Cancer &
337&
7.1\%&
149&
7.3\%\\
Cardiopulmonary &
508&
10.7\%&
218&
10.7\%\\
Cardiovascular &
1260&
26.6\%&
472&
23.1\%\\
Medical &
2021&
42.7\%&
964&
47.2\%\\
Neurology &
545&
11.5\%&
222&
10.9\%
 \\
Obstetrics &
9&
0.2\%&
4&
0.2\%
 \\
Psychiatric &
 &
&
&
\\
Other&
53&
1.1\%&
13&
0.6\%
\\ \midrule
Number of Previous Hospitalizations&&&&\\ \cmidrule(r){1-1}
0&
3622&
76.5\%&
1471&
72.0\%
\\
1&
643&
13.6\%&
302&
14.8\%
\\
2-5&
398&
8.4\%&
221&
10.8\%
\\
6+&
70&
1.5\%&
48&
2.4\%
\\ \midrule
Discharge Disposition&&&&\\ \cmidrule(r){1-1}
Expired&
419&
8.9\%&
159&
7.8\%
\\
Home&
2424&
51.2\%&
1027&
50.3\%
\\
Other&
120&
2.5\%&
70&
3.4\%
\\
Other Healthcare Facility&
341&
7.2\%&
219&
10.7\%
\\
Rehabilitation&
649&
13.7\%&
189&
9.3\%
\\
Skilled Nursing Facility&
780&
16.5\%&
378&
18.5\%
\\ \midrule
Binary Label Prevalence&&&&\\ \cmidrule(r){1-1}
Mortality&
419&
8.9\%&
159&
7.8\%
\\
Acute Kidney Injury (AKI)&
1070&
22.6\%
 & & \\ \midrule
Multilabel Prevalence (ICD9 Groups)&&&&\\ \cmidrule(r){1-1}
1:Infectious and
Parasitic Diseases &
1347&
28.5\%&
647&
31.7\%
\\
2:Neoplasms&
806&
17.0\%&
407&
19.9\%
\\
3:Endocrine, Nutritional and Metabolic Diseases, Immunity&
3288&
69.5\%&
1593&
78.0\%
\\
4:Blood and
Blood-Forming Organs&
1837&
38.8\%&
965&
47.3\%
\\
5:Mental Disorders&
1513&
32.0\%&
854&
41.8\%
\\
6:Nervous System
and Sense Organs&
1408&
29.8\%&
878&
43.0\%
\\
7:Circulatory System&
4026&
85.0\%&
1764&
86.4\%
\\
8:Respiratory System&
2338&
49.4\%&
1036&
50.7\%
\\
9:Digestive System&
1964&
41.5\%&
985&
48.2\%
\\
10:Genitourinary System&
2061&
43.6\%&
1069&
52.4\%
\\
11:Complications of Pregnancy, Childbirth,
and the Puerperium&
12&
0.3\%&
4&
0.2\%
\\
12:Skin and
Subcutaneous Tissue&
574&
12.1\%&
285&
14.0\%
\\
13:Musculoskeletal System and Connective Tissue&
1009&
21.3\%&
594&
29.1\%
\\
14:Congenital Anomalies&
157&
3.3\%&
87&
4.3\%
\\
15:Symptoms&
1385&
29.3\%&
786&
38.5\%
\\
16:Nonspecific Abnormal Findings&
421&
8.9\%&
250&
12.2\%
\\
17:Ill-defined and Unknown Causes of Morbidity and Mortality&
156&
3.3\%&
115&
5.6\%
\\
18:Injury and Poisoning&
2047&
43.3\%&
891&
43.6\%
\\
19:Supplemental V-Codes&
2475&
52.3\%&
1391&
68.1\%
\\
20:Supplemental E-Codes&
1517&
32.1\%&
869&
42.6\%
\\
  \end{longtabu}

  \begin{longtabu}{p{1cm}p{6cm}XXX}\caption{List of input features used in the model.} \label{tab:features} \\ \toprule
Index&
Observation Name&
LOINC code&
MIMIC specific code&
Units\\\midrule
\endhead
\bottomrule
\endfoot
0&
Heart Rate&
&
211&
bpm\\
1&
SpO2&
&
646&
percent\\
2&
Respiratory Rate&
&
618&
bpm\\
3&
Heart Rate&
&
220045&
bpm\\
4&
Respiratory Rate&
&
220210&
breaths per min\\
5&
O2 saturation pulseoxymetry&
&
220277&
percent\\
6&
Arterial BP [Systolic]&
&
51&
mmhg\\
7&
Arterial BP [Diastolic]&
&
8368&
mmhg\\
8&
Arterial BP Mean&
&
52&
mmhg\\
9&
Urine Out Foley&
&
40055&
ml\\
10&
HR Alarm [High]&
&
8549&
bpm\\
11&
HR Alarm [Low]&
&
5815&
bpm\\
12&
SpO2 Alarm [Low]&
&
5820&
percent\\
13&
SpO2 Alarm [High]&
&
8554&
percent\\
14&
Resp Alarm [High]&
&
8553&
bpm\\
15&
Resp Alarm [Low]&
&
5819&
bpm\\
16&
SaO2&
&
834&
percent\\
17&
HR Alarm [Low]&
&
3450&
bpm\\
18&
HR Alarm [High]&
&
8518&
bpm\\
19&
Resp Rate&
&
3603&
breaths\\
20&
SaO2 Alarm [Low]&
&
3609&
cm h2o\\
21&
SaO2 Alarm [High]&
&
8532&
cm h2o\\
22&
Previous WeightF&
&
581&
kg\\
23&
NBP [Systolic]&
&
455&
mmhg\\
24&
NBP [Diastolic]&
&
8441&
mmhg\\
25&
NBP Mean&
&
456&
mmhg\\
26&
NBP Alarm [Low]&
&
5817&
mmhg\\
27&
NBP Alarm [High]&
&
8551&
mmhg\\
28&
Non Invasive Blood Pressure mean&
&
220181&
mmhg\\
29&
Non Invasive Blood Pressure systolic&
&
220179&
mmhg\\
30&
Non Invasive Blood Pressure diastolic&
&
220180&
mmhg\\
31&
Foley&
&
226559&
ml\\
32&
CVP&
&
113&
mmhg\\
33&
Arterial Blood Pressure mean&
&
220052&
mmhg\\
34&
Arterial Blood Pressure systolic&
&
220050&
mmhg\\
35&
Arterial Blood Pressure diastolic&
&
220051&
mmhg\\
36&
ABP Alarm [Low]&
&
5813&
mmhg\\
37&
ABP Alarm [High]&
&
8547&
mmhg\\
38&
GCS Total&
&
198&
missing\\
39&
Hematocrit&
4544-3&
&
percent\\
40&
Potassium&
2823-3&
&
meq per l\\
41&
Hemoglobin [Mass/volume] in Blood&
718-7&
&
g per dl\\
42&
Sodium&
2951-2&
&
meq per l\\
43&
Creatinine&
2160-0&
&
mg per dl\\
44&
Chloride&
2075-0&
&
meq per l\\
45&
Urea Nitrogen&
3094-0&
&
mg per dl\\
46&
Bicarbonate&
1963-8&
&
meq per l\\
47&
Platelet Count&
777-3&
&
k per ul\\
48&
Anion Gap&
1863-0&
&
meq per l\\
49&
Temperature F&
&
678&
deg f\\
50&
Temperature C (calc)&
&
677&
deg f\\
51&
Leukocytes [\#/volume] in Blood by Manual count&
804-5&
&
k per ul\\
52&
Glucose&
2345-7&
&
mg per dl\\
53&
Erythrocyte mean corpuscular hemoglobin concentration [Mass/volume] by Automated count&
786-4&
&
percent\\
54&
Erythrocyte mean corpuscular hemoglobin [Entitic mass] by Automated count&
785-6&
&
pg\\
55&
Erythrocytes [\#/volume] in Blood by Automated count&
789-8&
&
per nl\\
56&
Erythrocyte mean corpuscular volume [Entitic volume] by Automated count&
787-2&
&
fl\\
57&
Erythrocyte distribution width [Ratio] by Automated count&
788-0&
&
percent\\
58&
Temp/Iso/Warmer [Temperature degrees C]&
&
8537&
deg f\\
59&
FIO2&
&
3420&
percent\\
60&
Magnesium&
2601-3&
&
mg per dl\\
61&
CVP Alarm [High]&
&
8548&
mmhg\\
62&
CVP Alarm [Low]&
&
5814&
mmhg\\
63&
Calcium [Moles/volume] in Serum or Plasma&
2000-8&
&
mg per dl\\
64&
Phosphate&
2777-1&
&
mg per dl\\
65&
FiO2 Set&
&
190&
torr\\
66&
Temp Skin [C]&
&
3655&
in\\
67&
pH&
11558-4&
&
u\\
68&
Temperature Fahrenheit&
&
223761&
deg f\\
69&
Central Venous Pressure&
&
220074&
mmhg\\
70&
Inspired O2 Fraction&
&
223835&
percent\\
71&
PAP [Systolic]&
&
492&
mmhg\\
72&
PAP [Diastolic]&
&
8448&
mmhg\\
73&
Calculated Total CO2&
34728-6&
&
meq per l\\
74&
Oxygen [Partial pressure] in Blood&
11556-8&
&
mmhg\\
75&
Base Excess&
11555-0&
&
meq per l\\
76&
Carbon dioxide [Partial pressure] in Blood&
11557-6&
&
mmhg\\
77&
PTT&
3173-2&
&
s\\
78&
Deprecated INR in Platelet poor plasma by Coagulation assay&
5895-7&
&
ratio\\
79&
PT&
5902-2&
&
s\\
80&
Temp Axillary [F]&
&
3652&
deg f\\
81&
Day of Life&
&
3386&
\\
82&
Total Fluids cc/kg/d&
&
3664&
\\
83&
Present Weight (kg)&
&
3580&
kg\\
84&
Present Weight (lb)&
&
3581&
cm h2o\\
85&
Present Weight (oz)&
&
3582&
cm h2o\\
86&
Fingerstick Glucose&
&
807&
mg per dl\\
87&
PEEP set&
&
220339&
cm h2o\\
88&
Previous Weight (kg)&
&
3583&
kg\\
89&
Weight Change (gms)&
&
3692&
g\\
90&
PEEP Set&
&
506&
cm h2o\\
91&
Mean Airway Pressure&
&
224697&
cm h2o\\
92&
Tidal Volume (observed)&
&
224685&
ml\\
93&
Resp Rate (Total)&
&
615&
bpm\\
94&
Minute Volume Alarm - High&
&
220293&
l per min\\
95&
Minute Volume Alarm - Low&
&
220292&
l per min\\
96&
Apnea Interval&
&
223876&
s\\
97&
Minute Volume&
&
224687&
l per min\\
98&
Paw High&
&
223873&
cm h2o\\
99&
Peak Insp. Pressure&
&
224695&
cm h2o
\end{longtabu}

\end{document}